\documentclass[12pt]{article}
\usepackage{amsmath}
\usepackage{amsfonts}
\usepackage{amssymb}
\usepackage{graphicx}
\usepackage{algorithmic}
\usepackage{algorithm}
\providecommand{\U}[1]{\protect \rule{.1in}{.1in}}
\providecommand{\U}[1]{\protect \rule{.1in}{.1in}}

\begin{document}

\title{FI-CBL: A Probabilistic Method for Concept-Based Learning with Expert Rules}
\author{Lev V. Utkin, Andrei V. Konstantinov and Stanislav R. Kirpichenko\\{\small {Higher School of Artificial Intelligence Technologies} }\\{\small {Peter the Great St.Petersburg Polytechnic University}}\\{\small {St.Petersburg, Russia} }\\{\small {e-mail: lev.utkin@gmail.com, andrue.konst@gmail.com},
kirpichenko.sr@gmail.com}}
\date{}
\maketitle

\begin{abstract}
A method for solving concept-based learning (CBL) problem is proposed. The
main idea behind the method is to divide each concept-annotated image into
patches, to transform the patches into embeddings by using an autoencoder, and
to cluster the embeddings assuming that each cluster will mainly contain embeddings
of patches with certain concepts. To find concepts of a new image, the method
implements the frequentist inference by computing prior and posterior
probabilities of concepts based on rates of patches from images with certain
values of the concepts. Therefore, the proposed method is called the
Frequentist Inference CBL (FI-CBL). FI-CBL allows us to incorporate the expert rules in the form of logic functions into the inference procedure. An idea behind the incorporation is to update prior and conditional probabilities of concepts to satisfy the rules. The method is transparent
because it has an explicit sequence of probabilistic calculations and a clear
frequency interpretation. Numerical experiments show that FI-CBL outperforms
the concept bottleneck model in cases when the number of training data is
small. The code of proposed algorithms is publicly available.


\end{abstract}

\section{Introduction}

Concept-based machine learning (CBL) is an innovative and promising approach
that focuses on utilizing high-level concepts derived from raw features to
express predictions in machine learning models, rather than using the raw
features themselves \cite{lage2020learning}. This approach aims to integrate
expert knowledge or human-like reasoning into machine learning models, leading
to more efficient and accurate predictions. By incorporating high-level
concepts, CBL can significantly improve explainability of the machine learning
model outputs, making them more accessible to users
\cite{Gupta-Narayanan-24,kim2018interpretability,wang2023learning,yeh2020completeness}%
.

One of the key types of CBL is the concept-based bottleneck model (CBM) which
can be regarded as a technique used for learning high-level representations of
data by forcing the model to learn a compressed and low-dimensional
representation of input features \cite{koh2020concept}. This low-dimensional
representation is referred to as the bottleneck. In the context of CBL, the
CBM learns high-level concepts by transforming the raw features into a
low-dimensional space, which allows the model to capture the essential
features while discarding irrelevant information. At that, the classifier
deriving final labels has access only to the concept representation, and the
decision is strongly tied to the concepts \cite{wang2023learning}. CBMs have
been widely used in various machine learning tasks, such as image recognition,
natural language processing, and speech recognition \cite{poeta2023concept}.
The CBM effectiveness lies in their ability to learn meaningful high-level
concepts, which can be easily interpreted and understood by humans. This
capability makes CBMs a powerful tool for developing explainable and
interpretable machine learning models.

It is important to points out that most models implementing CBL are based on
applying a deep neural network which transform raw features or images into a
specific low-dimensional representation, for example, the bottleneck, which
contains information about the concept values. As a result, the neural network
may be complex and require a huge amount of labeled instances for training,
which are not available in some cases. Therefore, we propose an extremely
simple and transparent model for CBL which is motivated by the work
\cite{Yamamoto-etal-2019}, devoted to annotating the histopathology images,
and by an interesting observation of the relationship between multiple
instance models (MIL)
\cite{Amores-13,Carbonneau-etal-18,Yao-Zhu-etal-20,Zhou-04}. MIL is a type of
weakly supervised learning, which deals with two concepts: bags and instances.
Each bag is labeled, and it consists of many instances or some its elements.
For example, a histology digital image obtained from the glass microscope
slides with a label indicating a disease, for example, cancer or non-cancer,
can be viewed as a \textquotedblleft bag\textquotedblright \ consisting of
patches extracted from the image, which are referred to as \textquotedblleft
instances\textquotedblright.

It turns out that the MIL can be regarded as a special case of the
concept-based learning where labels (concepts) of instances are unknown, but
there is a label description of the whole image (the bag). Yamamoto et al.
\cite{Yamamoto-etal-2019} proposed a simple algorithm for annotating instances
(patches) by having annotated the whole images. The algorithm is based on
clustering of the patch embeddings and computing probabilities that patches in
each cluster are malignant or benign in a simplest way by calculating rates of
patches from the malignant and benign images corresponding to the patches.

We extend the algorithm proposed by Yamamoto et al. \cite{Yamamoto-etal-2019}
to CBL and develop a simple method for determining concepts of new images. The
method is based on the frequentist inference and on computing prior and
posterior probabilities of concepts using rates of patches from images with
certain values of concepts. In other words, we calculate the relative
frequencies of patches from images with the concepts. Therefore, the proposed
method is called the Frequentist Inference CBL (FI-CBL). In addition, we
propose approaches to incorporate the knowledge-based expert rules of the form
\textquotedblleft IF ..., THEN ...\textquotedblright, which are elicited from
experts and constructed by means of concepts. For example, the rule from the
lung cancer diagnostics of a nodule can look like \textquotedblleft IF
\textbf{Contour} is
$<$%
spicules%
$>$%
, \textbf{Inclusion} is
$<$%
necrosis%
$>$%
, THEN a \textbf{Diagnosis} is
$<$%
malignant%
$>$%
. Here concepts are shown in Bold, their values are in angle brackets. It is
important to note that the approach to incorporate the knowledge-based expert
rules into neural networks in the framework of CBL has been proposed in
\cite{Konstantinov-Utkin-24}. Its idea is to add a special layer to a
classification neural network, which computes a probability distribution of
concepts in accordance with the available expert rules. According to
\cite{Konstantinov-Utkin-24}, the probability distributions of concepts are
approximately generated by the neural network and are corrected by the
incorporated expert rules. We propose another approach for taking into account
the expert rules. It is based on the combination of the Bayes rule and the
multinomial distribution. The expert rules in the form of logic functions
update prior probabilities of concepts as well as conditional probabilities in
the Bayes rule. This is a simple way for using the expert rules as a specific
type of constraints on probabilities of concepts. We have to note that the
term \textquotedblleft expert rules\textquotedblright \ is used in the proposed
model in a broader sense as an arbitrary logical function of concepts.

The code of proposed algorithms is available in:

https://github.com/NTAILab/simple\_concepts.

\section{Related work}

\textbf{Concept-based learning models.} Starting from the works
\cite{kim2018interpretability,yeh2020completeness}, many CBL models have been
proposed to implement ideas behind the concept-based learning under various
conditions. One of the goals to develop the CBL approaches is to interpret and
explain predictions of machine learning models. In order to achieve this goal
and to overcome some difficulties of the interpretation, a CBL model was
proposed in \cite{lage2020learning}. Concepts in this model are fully
transparent, thus enabling users to validate whether mental and machine models
are aligned. A concept-based explanation framework was presented in
\cite{dreyer2023understanding}. An algorithm for learning visual concepts from
images by applying a Bayesian generalization model was developed in
\cite{jia2013visual}. It should be noted that not only images (visual data)
are used in the CBL models, but also tabular data. For example, the concept
attribution approach to tabular learning and a definition of concepts over
tabular data were proposed in \cite{pendyala2022concept}. Applications of CBL
goes beyond the explanation and extends to a wide variety of problems.
Approaches for analysis time-series data using the CBL models were presented
in \cite{obermair2023example,tang2020interpretable}. The well-known anomaly
detection task in the framework of CBL was considered in
\cite{choi2023concept,sevyeri2023transparent}. One of the important areas of
the CBL application is medicine where doctors prefer explanations that are
user-friendly and represented via natural language \cite{Hendricks-etal-2018}.
Several authors have contributed into development of the medicine CBL models
\cite{marcinkevivcs2024interpretable,Meldo-Utkin-etal-2020,patricio2023coherent,vats2023concept,yan2023robust}%
. Survey papers \cite{Gupta-Narayanan-24,lee2023neural,mahinpei2021promises}
comprehensively discuss many aspects of the CBL models and their applications.

\textbf{Concept bottleneck models.} Most CBL models are implemented in the
form of the CBMs \cite{koh2020concept}. Due to the efficient and transparent
two-module architecture of CBMs, where the first module (a neural network)
implements the dependence of concepts on input instances, and the second
module implements the dependence of the target variable on the concepts, many
modifications and extensions of these models have been proposed
\cite{kim2018interpretability,Sheth-Kahou-23,yuksekgonul2022post}.

An extension of CBMs is the concept embedding model \cite{espinosa2022concept}
which learns two embeddings per concept, one for when it is active, and
another when it is inactive. The model aims to overcome the current
accuracy-vs-interpretability trade-off. Ideas behind the concept embedding
model have been used in the concept bottleneck generative models
\cite{ismail2023concept}.

Similarly to CBL, extensions and modifications of CBMs were motivated by their
applications or different conditions of their applications. For example, to
model the ambiguity in the concept predictions, a probabilistic CBM was
introduced in \cite{kim2023probabilistic}. Conditions of independence across
concepts were studied in \cite{Raman-etal-24}. Different aspects of the
concept-based interventions were considered in \cite{Marcinkevics-etal-24}. An
application of CBMs to the images segmentation and tracking was presented in
\cite{pittino2023hierarchical}. Two causes of performance disparity between
soft (inputs to a label predictor are the concept probabilities) and hard (the
label predictor only accepts binary concepts) CBMs were proposed in
\cite{havasi2022addressing}. The CLIP-based CBMs using the well-known CLIP
model \cite{radford2021learning} was proposed in \cite{kazmierczak2023clip}.

The above modifications and extensions of CBMs can be regarded as a small part
of all available modifications
\cite{chauhan2023interactive,cui2023ceir,marconato2022glancenets,margeloiu2021concept,Sun-Yan-etal-24}%
, which are caused by the great interest in CBL.

\textbf{Incorporating expert rules into machine learning models}. The idea of
combining the prior expert knowledge with machine learning models has already
attracted some interest, and a number of interesting approaches have been
proposed for its implementation. A comprehensive and exhaustive review of
various available approaches to integrate prior knowledge into the training
process was presented in \cite{von2021informed}. Authors in
\cite{von2021informed} also propose a concept of informed machine learning,
which can be viewed as a uniting term for different approaches.

Depending on the knowledge representation, there are different approaches for
the knowledge integration into the machine learning pipeline. We do not touch
upon a large class of methods related to the representation of knowledge in
forms of algebraic equations, differential equations, probabilistic relations,
etc. An analysis and review of these methods can again be found in
\cite{von2021informed}. Our goal is to study the knowledge representation in
the form of expert rules or logic rules. A common approach to incorporate the
logic rules into a machine learning model is to add the rules as constraints
to loss functions
\cite{diligenti2017semantic,diligenti2017integrating,hu2016deep,xu2018semantic}%
. However, this approach does not guarantee that the rules will be satisfied
for all training and testing instances because violation of the constraints is
only penalized, but not eliminated. Another way to integrate the rules into
neural networks is to map components of the rules to neurons
\cite{francca2014fast,garcez2019neural}. In this approach, a neural network
implements logic functions corresponding to the expert rules. An interesting
approach has been proposed in \cite{Yang-etal-abductive-24} where authors
present an effective safe abductive learning method and show that induction
and abduction are mutually beneficial. An extensive review of methods for
updating models based on expert feedback is presented in
\cite{chen2023perspectives}.

A quite different approach to incorporate expert rules into machine learning
in the framework of CBL was presented in \cite{Konstantinov-Utkin-24}.
Following this approach, we present a computationally simple algorithm for
predicting the concept probabilities and provide a way to incorporate the
expert rules in the form of logic functions.

\section{Background}

A common statement of the CBL problem is based on considering a classifier
which predicts a set of concepts as well as the target variable
\cite{xu2023statistically}.

Suppose that a training set is represented as a set of triples $(\mathbf{x}%
_{i},y_{i},\mathbf{c}_{i})$, $i=1,...,N$, where $\mathbf{x}_{i}\in$
$\mathcal{X}\subset \mathbb{R}^{d}$ is the input feature vector; $y_{i}%
\in \mathcal{Y}=\{1,2,...,K\}$ is the corresponding target defining $K$-class
classification task; $\mathbf{c}_{i}=(c_{i}^{(1)},...,c_{i}^{(m)}%
)\in \mathcal{C}$ is a set of $m$ concepts $\mathbf{c}_{i}=(c_{i}%
^{(1)},...,c_{i}^{(m)})\in \mathcal{C}$ which are given with targets. In most
works, concepts are represented as a vector $\mathbf{c}_{i}$ with $m$ binary
elements such that $c_{i}^{(j)}=1$ denotes that the $j$-th concept is present
in a description of the input $\mathbf{x}_{i}$, and $c_{i}^{(j)}=0$ denotes
that the $j$-th concept is not present.

One of the CBL goals is to predict targets and concepts that is to find the
dependence $h:\mathcal{X}\rightarrow(\mathcal{C},\mathcal{Y})$ on concepts and
inputs. Another goal is to explain what concepts of the input are responsible
for the corresponding prediction. In other words, the CBL model aims to
interpret how predictions depend on concepts of the corresponding inputs. The
above goals can be achieved by applying CBM proposed by Koh et al.
\cite{koh2020concept} as an important type of the CBL models. The function $h$
in the CBM is represented as two functions: the first one $g:$ $\mathcal{X}%
\rightarrow \mathcal{C}$ maps the input vector to concepts; the second function
$f:\mathcal{C}\rightarrow \mathcal{Y}$ maps the concepts to the outputs. The
prediction $y$ for a new instance $\mathbf{x}$ can be obtained as
$y=f(g(\mathbf{x}))$. Here concepts act as a bottleneck in the interpretation
of predictions.

\section{The model and its training\label{sec:model_and_its_training}}

It is assumed that each concept $c^{(i)}$ can take a value from the set
$\mathcal{C}^{(i)}=\{1,...,n_{i}\}$ called the $i$-th concept outcome set,
$i\in \{0,\dots,m\}$. The concept $c^{(0)}$ is a special concept corresponding
to the target variable $y$.

We also suppose that there are $N$ images $\mathbf{x}_{i}$, $i=1,...,N$, in
the training set such that the $i$-th image is characterized by a set of
concept values $\mathbf{c}_{i}=(c_{i}^{(0)},...,c_{i}^{(m)})$. The whole
dataset consists of pairs $(\mathbf{x}_{i},\mathbf{c}_{i})$ of vectors. For
example, the lung cancer nodule description can be based on two concepts:
\textbf{Contour }$c^{(1)}$ and \textbf{Inclusion }$c^{(2)}$. The concept
\textbf{Contour} takes values
$<$%
smooth%
$>$%
\
$<$%
grainy%
$>$%
,
$<$%
spicules%
$>$%
, the concept \textbf{Inclusion} takes values
$<$%
homogeneous%
$>$%
\ and
$<$%
necrosis%
$>$%
. The target value $y$ or the concept \textbf{Diagnosis} $c^{(0)}$ takes
values:
$<$%
malignant%
$>$%
\ and
$<$%
benign%
$>$%
. Then we have the formal concept description $\mathcal{C}^{(0)}=\{1,2\}$,
$\mathcal{C}^{(1)}=\{1,2,3\}$, $\mathcal{C}^{(2)}=\{1,2\}$.

Let us divide each image $\mathbf{x}_{i}$ into $s$ patches of the same
dimension denoted as $\xi_{1}^{(i)},...,\xi_{s}^{(i)}$. By using an
autoencoder, we can obtain the corresponding embeddings $e_{1}^{(i)}%
,...,e_{s}^{(i)}$ of a smaller dimension.

In fact, we have a weakly supervised learning task where labels of the whole
images are known, including values of all concepts or a part of concepts, but
labels of patches are unknown. However, we can compute probabilities of labels
for patches by separating embeddings corresponding to patches into groups
(clusters) with different contents and by counting up how many whole images
having a certain concept value contain patches from each group (cluster).

All embeddings are clustered into $R$ clusters $K_{1},...,K_{R}$, i.e., we
obtain subsets of embeddings, which fall into $k$-th cluster, of the form:
\begin{equation}
\left \{  e_{j}^{(i)},i\in \mathcal{I}_{k},j\in \mathcal{J}_{k}\right \}  ,
\end{equation}
where $\mathcal{I}_{k}$ and $\mathcal{J}_{k}$ are index sets such that the
$k$-th cluster contains embeddings of patches with indices from $\mathcal{J}%
_{k}$ belonging to indices of images from $\mathcal{I}_{k}$.

Clusters contain $s_{1},...,s_{R}$ embeddings such that $s_{1}+...+s_{R}=S$,
where $S$ is the total number of patches obtained from all images. If all
images are divided into the same number of patches, then $S=s\cdot N$. Let us
write all concepts in the form of one vector consisting of concatenated
vectors of indices:%
\begin{equation}
\mathcal{C}=(\underbrace{1,...,n_{0}}_{\mathcal{C}^{(0)}},\underbrace
{1,...,n_{1}}_{\mathcal{C}^{(1)}},...,\underbrace{1,...,n_{m}}_{\mathcal{C}%
^{(m)}}).
\end{equation}

\begin{figure}
[ptb]
\begin{center}
\includegraphics[
height=2.5516in,
width=4.3265in
]%
{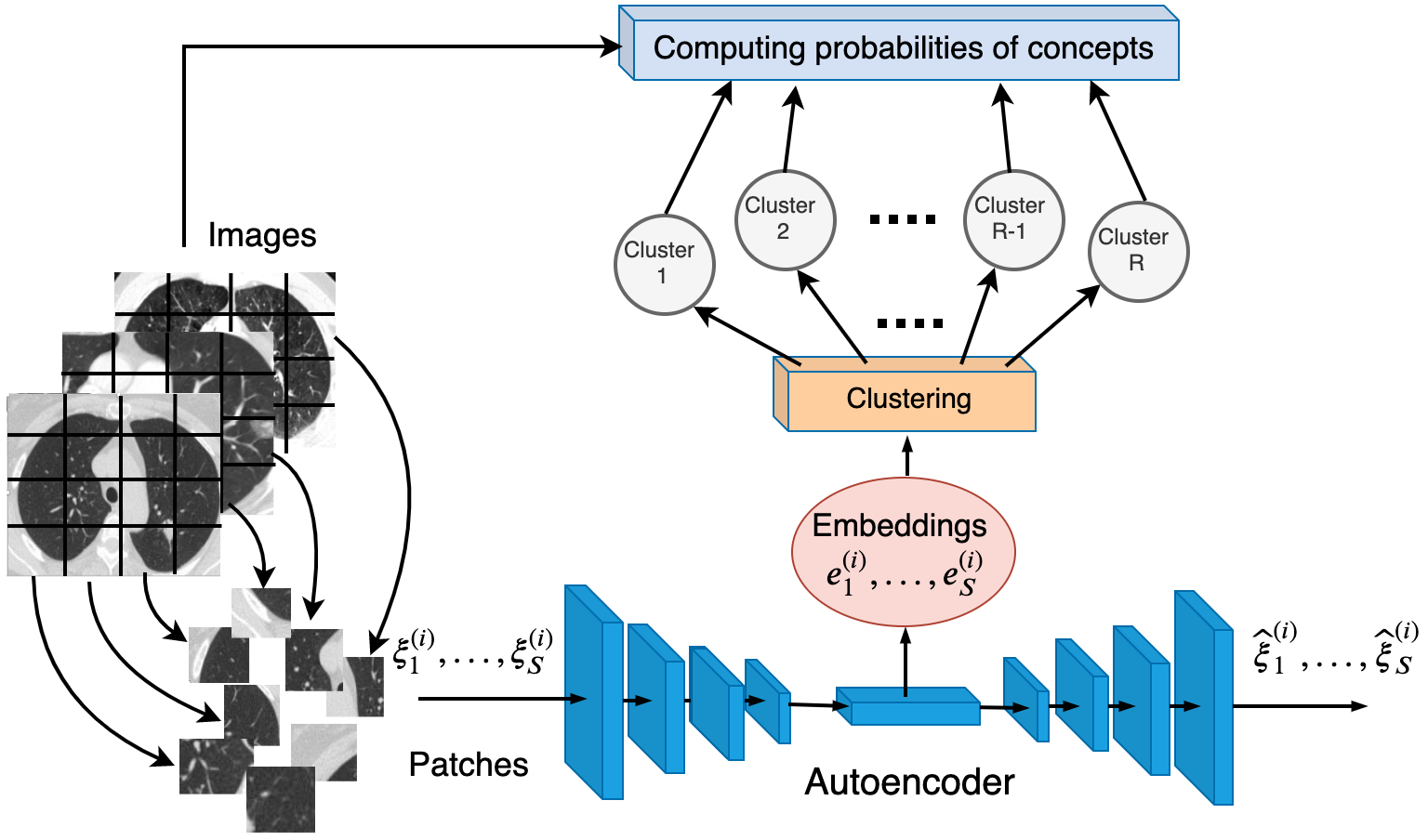}%
\caption{A general scheme of the image transformation to sets of clustered
embeddings of patches: each image is divided into $s$ patches $\xi_{j}^{(i)}$;
every patch is transformed into an embedding $e_{j}^{(i)}$; embeddings are
clustered into $R$ clusters; probabilities of concepts are computed by having
the image concept labels $\mathbf{c}_{i}$ and distributions of embeddings with
corresponding concept values in clusters}%
\label{f:general_scheme}%
\end{center}
\end{figure}

A general scheme of the above representation of the images in the form
embeddings of patches divided into $R$ clusters is depicted in Fig.
\ref{f:general_scheme}.%

If we assume that each concept is a random variable $C^{(r)}$ taking values
from $\mathcal{C}^{(r)}$, then we aim to find the conditional probability
$p(r,v\mid l)=P\left \{  C^{(r)}=v\mid e\in K_{l}\right \}  $ that the concept
$C^{(r)}$ takes the value $v$ under condition that the embedding $e$ of a
patch taken from a considered image falls into the cluster $K_{l}$. Let us
define the following additional probabilities and their short notations:

\begin{itemize}
\item $p(l\mid r,v)=P\left \{  e\in K_{l}\mid C^{(r)}=v\right \}  $ is the
conditional probability that an embedding in the cluster $K_{l}$ is from the
image having the value $v$ of the $r$-th concept;

\item $p(r,v)=P\left \{  C^{(r)}=v\right \}  $ is the prior probability that an
embedding is from the image having the value $v$ of the $r$-th concept;

\item $p(l)=P\left \{  e\in K_{l}\right \}  $ is the unconditional probability
that an embedding falls into the cluster $K_{l}$.
\end{itemize}

By using the Bayes rule, we write
\begin{equation}
p(r,v\mid l)=\frac{p(l\mid r,v)\cdot p(r,v)}{p(l)}. \label{Bayes_rule}%
\end{equation}

It is important to point out that we do not need probabilities $p(r,v\mid l)$
for the inference when a new instance is classified. We need to know only
$p(l\mid r,v)$ and $p(r,v)$ for the inference. On the other hand, the
probability $p(r,v\mid l)$ can be regarded as a measure for determining what
values of concepts embeddings in the $l$-th cluster have. The large
probability $p(r,v\mid l)$ implies that the embeddings in $K_{l}$ have mainly
the concept value $c^{(r)}=v$. If we assume that clusters are homogeneous to
some extent, then embeddings contained in the cluster $K_{l}$ correspond to
patches from images with the concept $c^{(r)}=v$. This is an important
information because it allows us to highlight areas in the image with given concepts.

Let us introduce the following additional notations:

\begin{itemize}
\item $s_{v}^{(r)}(l)$ is the number of embeddings in the $l$-th cluster
obtained from images with $c^{(r)}=v$;

\item $s_{v}^{(r)}$ is the total number of embeddings in all clusters obtained
from images with $c^{(r)}=v$.
\end{itemize}

The conditional probability $p(l\mid r,v)$ is determined as the proportion of
embeddings from images with $c^{(r)}=v$ that fall into the cluster $K_{l}$ to
the entire set of embeddings from the images with $c^{(r)}=v$, i.e., there
holds $p(l\mid r,v)=s_{v}^{(r)}(l)/s_{v}^{(r)}$.

The prior probability $p(r,v)$ is determined as the proportion of images with
$c^{(r)}=v$ to all images in the dataset, i.e., there holds $p(r,v)=s_{v}%
^{(r)}/S$. The unconditional probability $p(l)$ can be computed from the
condition:
\begin{equation}
\sum_{v=1}^{n_{r}}p(r,v\mid l)=1.
\end{equation}

It can be also determined as the proportion of embeddings in the cluster
$K_{l}$ to all embeddings from all images, i.e., there holds $p(l)=s_{l}/S$.
Hence, the posterior probability is computed as $p(r,v\mid l)=s_{v}%
^{(r)}(l)/s_{l}$.

\section{The model inference}

Suppose we have a new instance $\mathbf{x}$ consisting of $s$ patches $\xi
_{1},...,\xi_{s}$ which are fed to the trained autoencoder in order to obtain
embeddings $e_{1},...,e_{s}$. These embeddings are distributed among clusters
$K_{1},...,K_{R}$ according to their distances to the cluster centers. Let
$s_{1},...,s_{R}$ be numbers of embeddings which fall into clusters
$K_{1},...,K_{R}$, respectively, such that $s_{1}+...+s_{R}=s$. Let us denote
the set of embeddings, produced by a new instance, which fall into $K_{l}$, as
$E_{l}^{\ast}=\{e^{\ast}(1,l),...,e^{\ast}(s_{l},l)\}$. We do not have any
information about concepts and their values which describe the instance.
However, we can compute the probability that $C^{(r)}$ is equal to $v$ for all
$r=1,...,m$, $v=1,...,n_{r}$, under condition that $s_{1},...,s_{R}$
embeddings fall into clusters $K_{1},...,K_{R}$, respectively, i.e., find the
conditional probability
\begin{equation}
p(r,v\mid E_{1:R}^{\ast})=P\left \{  C^{(r)}=v\mid E_{1:R}^{\ast}%
,...,E_{R}^{\ast}\right \}  .
\end{equation}
Here $E_{1:R}^{\ast}$ is a short notation of $E_{1}^{\ast},...,E_{R}^{\ast}$.
This probability depends on the probabilities that embeddings with the concept
$c^{(r)}=v$ fall into $K_{i}$, $l=1,...,R$. This probability can be estimated
by means of the probability $p(l\mid r,v)$ which has been considered above and
defined as the proportion of embeddings from images with $c^{(r)}=v$ that fall
into the cluster $K_{i}$ to the entire set of embeddings from the images with
$c^{(r)}=v$.

Then we can write the following Bayes rule:%
\begin{equation}
p(r,v\mid E_{1:R}^{\ast})=\frac{p(E_{1:R}^{\ast}\mid r,v)\cdot p(r,v)}%
{P(E_{1:R}^{\ast})}, \label{Bayes_rule_inderence}%
\end{equation}
where $p(E_{1:R}^{\ast}\mid r,v)=P\{E_{1:R}^{\ast}\mid C^{(r)}=v\}$ is the
conditional probability that $s_{1},...,s_{R}$ embeddings fall into clusters
$K_{1},...,K_{R}$, respectively, under condition that the image has the value
$v$ of the $r$-th concept; $P\left \{  E_{1:R}^{\ast}\right \}  $ is
unconditional probability that $s_{1},...,s_{R}$ embeddings fall into clusters
$K_{1},...,K_{R}$.

We propose to apply the multinomial distribution with probabilities $p(l\mid
r,v)$ of events in order to represent $P\{E_{1:R}^{\ast}\mid C^{(r)}=v\}$:
\begin{equation}
p(E_{1:R}^{\ast}\mid r,v)=\frac{s!}{s_{1}!\cdot \cdot \cdot s_{R}!}\prod
_{l=1}^{R}p^{s_{l}}(l\mid r,v). \label{multinom_d}%
\end{equation}

The probability $p(l\mid r,v)$ has been also defined in (\ref{Bayes_rule}).
The unconditional probability $P\left \{  E_{1:R}^{\ast}\right \}  $ can be
computed by using the condition:
\begin{equation}
\sum_{v=1}^{n_{r}}p(r,v\mid E_{1:R}^{\ast})=1. \label{cond_prob}%
\end{equation}

The main difficulty of using (\ref{multinom_d}) is that some probabilities
$p(l\mid r,v)$ may be $0$. In this case, the product (\ref{multinom_d}) is
also $0$. In order to overcome this problem, we propose to replace the
zero-valued probabilities with some small value $\epsilon>0$. In this case,
the probability $P(E_{1:R}^{\ast})$ can be computed only by means of
(\ref{cond_prob}).

In sum, for a new instance, we compute $n_{1}+...+n_{m}$ probabilities
$p(r,v\mid E_{1:R}^{\ast})$ considering all $v=1,...,n_{r}$ and $r=1,...,m$.

Let us introduce the thresholds $\gamma_{r}$ for all values of the $r$-th
concept. If $p(r,v\mid E_{1:R}^{\ast})\geq \gamma_{r}$, then the value $v$ of
the concept $c^{(r)}$ is assigned to the instance. After considering all the
threshold conditions, we get the concept-based description of the image.

\section{Incorporating expert rules}

Let us introduce the logical literal denoted as $[c^{(j)}=i]$, which takes the
value $1$, if the concept $c^{(j)}$ has the value $i$. A set of expert rules
can be represented as a logical expression $g(\mathbf{c})$ over literals
$[c^{(j)}=i]$ such that $g(\mathbf{c})=0$ means that the rule is FALSE,
$g(\mathbf{c})=1$ means that it is TRUE. One of the forms of rules provided by
experts is represented as \textquotedblleft IF $F$, THEN $G$\textquotedblright%
, where $F$ is the antecedent, $G$ is the consequent. This rule is expressed
through the logical function as $F\rightarrow G=\lnot F\vee G$, where symbols
$\rightarrow$ and $\lnot$ denote operations of implication and negation, respectively.

Expert rules change prior probabilities of concepts $p(r,v)$ as well as
conditional probabilities of concepts $p(l\mid r,v)$. Therefore, the next
question we need to answer is how to update these probabilities in order to
implement the model inference taking into account the rules.

\subsection{Expert rules and prior probabilities of concepts}

First, we consider how prior probabilities of concepts are changed by using
the expert rules. Let us return to the conditional probability $p(r,v\mid
E_{1:R}^{\ast})$ in (\ref{Bayes_rule_inderence}). One of the ways to take into
account the expert rules is to assume that the expert rules change prior
probabilities of concepts $p(r,v)$, i.e., we have to find the conditional
probabilities $P\left \{  C^{(r)}=v\mid g(\mathbf{C})=1\right \}  $. Here
$\mathbf{C}$ is the random vector taking values from combinations of the
concept values. There exist $n_{0}\cdot \cdot \cdot n_{m}$ different
combinations of the concept values which are formed from the Cartesian product
$\mathcal{C=C}^{(0)}\times....\times \mathcal{C}^{(m)}$ defined above. Let
$\mathbf{z}\in \mathcal{C}$ be one of the combinations. Each expert rule takes
values TRUE or FALSE ($1$ or $0$) after substituting $\mathbf{z}$ into the
logical function $g$. Let us use again the Bayes rule for determining the
posterior probability $P\left \{  C^{(r)}=v\mid g(\mathbf{C})=1\right \}  $:
\begin{align}
&  P\left \{  C^{(r)}=v\mid g(\mathbf{C})=1\right \} \nonumber \\
&  =\frac{P\left \{  g(\mathbf{C})=1\mid C^{(r)}=v\right \}  \cdot
p(r,v)}{P\left \{  g(\mathbf{C})=1\right \}  }. \label{Bayes_Expert_rule}%
\end{align}

In order to determine the probability $P\left \{  g(\mathbf{C})=1\right \}  $,
we consider all cases when the random vector takes values $\mathbf{z}$ from
the set $\mathcal{C}$. However, when the conditional probability $P\left \{
g(\mathbf{C})=1\mid C^{(r)}=v\right \}  $ is determined, the the set
$\mathcal{C}$ of all vectors $\mathbf{z}$ is restricted by the condition
$C^{(r)}=v$. A new restricted set denoted as $\mathcal{C}_{r,v}$ is formed by
replacing $\mathcal{C}^{(r)}$ in $\mathcal{C}$ with the single value $v$,
i.e., $\mathcal{C}^{(r)}=\{v\}$ and
\begin{equation}
\mathcal{C}_{r,v}=\mathcal{C}^{(0)}\times....\times \{v\} \times...\times
\mathcal{C}^{(m)}.
\end{equation}

By using the rule of total probability, we write:%
\begin{equation}
P\{g(\mathbf{C})=1\mid C^{(r)}=v\}=\sum_{\mathbf{z}\in \mathcal{C}_{r,v}%
}P\{g(\mathbf{C})=1\mid \mathbf{C}=\mathbf{z}\}P\{ \mathbf{C}=\mathbf{z}\}.
\label{prob_rule_for_all}%
\end{equation}

Since the function $g(\mathbf{C})$ takes values $0$ and $1$, then the
conditional probability is determined as follows:%
\begin{equation}
P\{g(\mathbf{C})=1\mid C^{(r)}=v\}=\sum_{\mathbf{z}\in \mathcal{C}_{r,v}%
}g(\mathbf{z})\cdot P\{ \mathbf{C}=\mathbf{z}\}. \label{total_prob_2}%
\end{equation}

The probability $P\{ \mathbf{C}=\mathbf{z}\}$ can be found by considering all
images which have the combination $\mathbf{z}\in \mathcal{C}_{r,v}$ of concept
values, i.e., the probability is equal to the proportion of the number of
images with concepts $\mathbf{z}\in \mathcal{C}_{r,v}$ to the number $N$ of all
images in the dataset.

In sum, for every combination of concepts $\mathbf{z}$ $\in \mathcal{C}_{r,v}$,
we check whether the concepts satisfy the expert rule $g(\mathbf{z})$ and then
compute $P\{C=\mathbf{z}\}$ for this combination. It is important to point out
that a combination $\mathbf{z}$ is not considered if there are no images with
the corresponding set of concept values in the dataset because $P\{
\mathbf{C}=\mathbf{z}\}=0$ in this case.

The unconditional probability $P\left \{  g(\mathbf{C})=1\right \}  $ can be
obtained from the condition
\begin{equation}
\sum_{v=1}^{n_{r}}P\left \{  C^{(r)}=v\mid g(\mathbf{C})=1\right \}  =1,
\end{equation}
and is equal to
\begin{align}
P\left \{  g(\mathbf{C})=1\right \}   &  =\sum_{v=1}^{n_{r}}P\left \{
g(\mathbf{C})=1\mid C^{(r)}=v\right \}  \cdot p(r,v)\nonumber \\
&  =\sum_{v=1}^{n_{r}}\sum_{\mathbf{z}\in \mathcal{C}_{r,v}}g(\mathbf{z}%
_{v}^{(r)})\cdot P\{ \mathbf{C}=\mathbf{z}\} \cdot p(r,v).
\end{align}

Hence, we rewrite the expression (\ref{Bayes_Expert_rule}) for computing
$P\left \{  C^{(r)}=v\mid g(\mathbf{C})=1\right \}  $ as follows:
\begin{equation}
P\left \{  C^{(r)}=v\mid g(\mathbf{C})=1\right \}  =U_{v}^{(r)}\cdot p(r,v),
\label{Bayes_Expert_rule_final}%
\end{equation}
where
\begin{equation}
U_{v}^{(r)}=\frac{\sum_{\mathbf{z}\in \mathcal{C}_{r,v}}g(\mathbf{z})\cdot P\{
\mathbf{C}=\mathbf{z}\}}{P\left \{  g(\mathbf{C})=1\right \}  }.
\label{ER_update}%
\end{equation}

The prior probability $p(r,v)$ of the concept is updated in accordance with
the rule $g$ by means of its multiplying by the updating coefficient
$U_{v}^{(r)}$.

Finally, we have obtained the updated probabilities $p(r,v)$ which are used in
(\ref{Bayes_rule_inderence}) for computing posterior marginal probabilities of
concepts $P\left \{  C^{(r)}=v\mid E_{1:R}^{\ast}\right \}  $.

\subsection{Expert rules and conditional probabilities}

In addition to changes of prior probabilities, it is necessary to consider how
conditional probabilities $P\left \{  e\in K_{l}\mid C^{(r)}=v\right \}  $ are
updated due to expert rules. Let us denote the conditional event $e\in
K_{l}\mid C^{(r)}=v$ as $E_{l}^{(r)}=v$. Then we have to find the probability
\begin{equation}
P\left \{  E_{l}^{(r)}=v\mid g(\mathbf{C})=1\right \}  =\frac{P\left \{
g(\mathbf{C})=1\mid E_{l}^{(r)}=v\right \}  \cdot P\left \{  E_{l}%
^{(r)}=v\right \}  }{P\left \{  g(\mathbf{C})=1\right \}  }.
\label{Cond_Bayes_rule}%
\end{equation}

Note that the probability $P\left \{  E_{l}^{(r)}=v\right \}  $ is nothing else,
but the conditional probability $p(l\mid r,v)$ which has been defined above in
Sec. \ref{sec:model_and_its_training}, i.e., $P\left \{  E_{l}^{(r)}=v\right \}
=s_{v}^{(r)}(l)/s_{v}^{(r)}$. Let $\mathcal{C}_{r,v}(l)\subseteq$
$\mathcal{C}_{r,v}$ be a subset of $\mathcal{C}_{r,v}$, which contains the
concept combinations of embeddings from the cluster $K_{l}$. Then the
conditional probability $P\left \{  g(\mathbf{C})=1\mid E_{l}^{(r)}=v\right \}
$ is determined as
\begin{align}
P\{g(\mathbf{C})  &  =1\mid E_{l}^{(r)}=v\} \nonumber \\
&  =\sum_{\mathbf{z}\in \mathcal{C}_{r,v}(l)}P\{g(\mathbf{C})=1\mid
\mathbf{C}=\mathbf{z}\}P\{ \mathbf{C}=\mathbf{z}\}.
\label{prob_rule_for_cluster}%
\end{align}

It can be seen from (\ref{prob_rule_for_cluster}) that $P\{g(\mathbf{C})=1\mid
E_{l}^{(r)}=v\}$ differs from $P\{g(\mathbf{C})=1\mid C_{l}^{(r)}=v\}$ in
(\ref{prob_rule_for_all}) in that the set $\mathcal{C}_{r,v}(l)$ is limited to
considering only the cluster $K_{l}$. Hence, there holds%
\begin{equation}
P\{g(\mathbf{C})=1\mid E_{l}^{(r)}=v\}=\sum_{\mathbf{z}\in \mathcal{C}%
_{r,v}(l)}g(\mathbf{z})\cdot P\{ \mathbf{C}=\mathbf{z}\}.
\label{cond_total_prob_2}%
\end{equation}

We rewrite the expression (\ref{Cond_Bayes_rule}) for computing $P\left \{
E_{l}^{(r)}=v\mid g(\mathbf{C})=1\right \}  $ as follows:%

\begin{align}
P\left \{  E_{l}^{(r)}=v\mid g(\mathbf{C})=1\right \}   &  =V_{v}^{(r)}(l)\cdot
P\left \{  E_{l}^{(r)}=v\right \} \nonumber \\
&  =p(l\mid r,v)\nonumber \\
&  =V_{v}^{(r)}(l)\cdot s_{v}^{(r)}(l)/s_{v}^{(r)}, \label{cond_total_prob_3}%
\end{align}
where
\begin{equation}
V_{v}^{(r)}(l)=\frac{\sum_{\mathbf{z}\in \mathcal{C}_{r,v}(l)}g(\mathbf{z}%
)\cdot P\{ \mathbf{C}=\mathbf{z}\}}{P\left \{  g(\mathbf{C})=1\right \}  }.
\end{equation}

Note that there holds
\[
\sum_{l=1}^{R}P\left \{  E_{l}^{(r)}=v\mid g(\mathbf{C})=1\right \}  =1,
\]
because the event $E_{l}^{(r)}=v$ means that the embedding $e$ falls into one
of the clusters. Hence, we can write%
\[
\sum_{l=1}^{S}V_{v}^{(r)}(l)\cdot p(l\mid r,v)=1.
\]

The conditional probability $P\left \{  E_{l}^{(r)}=v\mid g(\mathbf{C}%
)=1\right \}  $ is updated in accordance with the rule $g$ by means of its
multiplying by the updating coefficient $V_{v}^{(r)}(l)$.

After updating the probabilities, they are substituted into
(\ref{Bayes_rule_inderence}) and (\ref{multinom_d}) in order to obtain
probabilities of concepts for a new instance.

\subsection{Expert rules with uncertainty}

So far, we have considered the hard expert-convinced rules, i.e., the rules
taking the value TRUE with the unit probability. Let us study the case when
rules are of the form \textquotedblleft IF $F$, THEN $G$ with probability
$\pi$\textquotedblright. There are different interpretations of the
implication operation probabilities \cite{nguyen2002probability}. We apply an
interpretation which considers the probability of $P(\lnot F\vee G)$ that
either $G$ is true or $F$ is false.

A simple way to adapt the uncertain rules to the proposed scheme is to replace
the value $g(\mathbf{z})$ in (\ref{ER_update}) with probabilities $\pi$ and
$1-\pi$. Let us rewrite (\ref{total_prob_2}) taking into account the
probability $\pi(\mathbf{z})=P\{g(\mathbf{z})=1\}$ that the rule is TRUE for
$\mathbf{z}$, i.e., is satisfied for the combination $\mathbf{z}$ of concepts,
as follows:%
\begin{equation}
P\{g(\mathbf{C})=1\mid C^{(r)}=v\}=\sum_{\mathbf{z}\in \mathcal{C}_{r,v}}%
\pi(\mathbf{z})\cdot P\{ \mathbf{C}=\mathbf{z}\}.
\end{equation}

It can be seen from the above that only the updating coefficient $U_{v}^{(r)}$
in (\ref{ER_update}) is changed when the probability of the rule is added.

In the same way, the conditional probabilities $P\left \{  E_{l}^{(r)}=v\mid
g(\mathbf{C})=1\right \}  $ in (\ref{cond_total_prob_3}) can be updated. In the
case, we change only the updating coefficient $V_{v}^{(r)}(l)$ as follows:
\begin{equation}
V_{v}^{(r)}(l)=\frac{\sum_{\mathbf{z}\in \mathcal{C}_{r,v}(l)}\pi
(\mathbf{z})\cdot P\{ \mathbf{C}=\mathbf{z}\}}{P\left \{  g(\mathbf{C}%
)=1\right \}  }.
\end{equation}

It should be noted that the uncertainty representation of expert rules is a
special important question which requires a detailed separate  investigation.
Therefore, it is not considered in this work and will be study in future. 

\section{Illustrative example}

\subsection{Concepts}

\subsubsection{Training}

To illustrate calculation in FI-CBL, we consider an illustrative example with
concepts \textbf{Contour }$c^{(1)}$ and \textbf{Inclusion }$c^{(2)}$ of the
lung cancer nodule description given in Section
\ref{sec:model_and_its_training}. The vector of all concept values is
\begin{equation}
\mathcal{C}=(\underbrace{1,2}_{\mathcal{C}^{(0)}},\underbrace{1,2,3}%
_{\mathcal{C}^{(1)}},\underbrace{1,2}_{\mathcal{C}^{(2)}}).
\end{equation}
%

\begin{table}[tbp] \centering
\caption{Concept values for the illustrative example with the lung nodules}%
\begin{tabular}
[c]{cccc}\hline
\textbf{Diagnosis} & \textbf{Contour} & \textbf{Inclusion} & \# images\\ \hline
benign & smooth & homogeneous & $2$\\
benign & grainy & homogeneous & $2$\\
malignant & spicules & homogeneous & $4$\\
malignant & grainy & necrosis & $2$\\ \hline
\end{tabular}
\label{t:example_descr_1}%
\end{table}%

Suppose we have $10$ images whose description is shown in Table
\ref{t:example_descr_1}, where the first three columns contain values of the
concepts and the fourth column indicates the number of images having the
corresponding concept values. The images are divided into $4$ patches. Numbers
of images with values
$<$%
malignant%
$>$%
\ and
$<$%
benign%
$>$%
\ are $6$ and $4$, respectively. Suppose that embeddings corresponding to
patches are clustered to $R=3$ clusters such that:

\begin{itemize}
\item the first cluster contains embeddings of the background patches;

\item the second cluster contains embeddings of nodules with spicules or with
the grainy contour and with necrosis;

\item the third cluster contains embeddings of nodules which are smooth or
grainy and homogeneous.
\end{itemize}

\begin{figure}
[ptb]
\begin{center}
\includegraphics[
height=3.0294in,
width=3.5795in
]%
{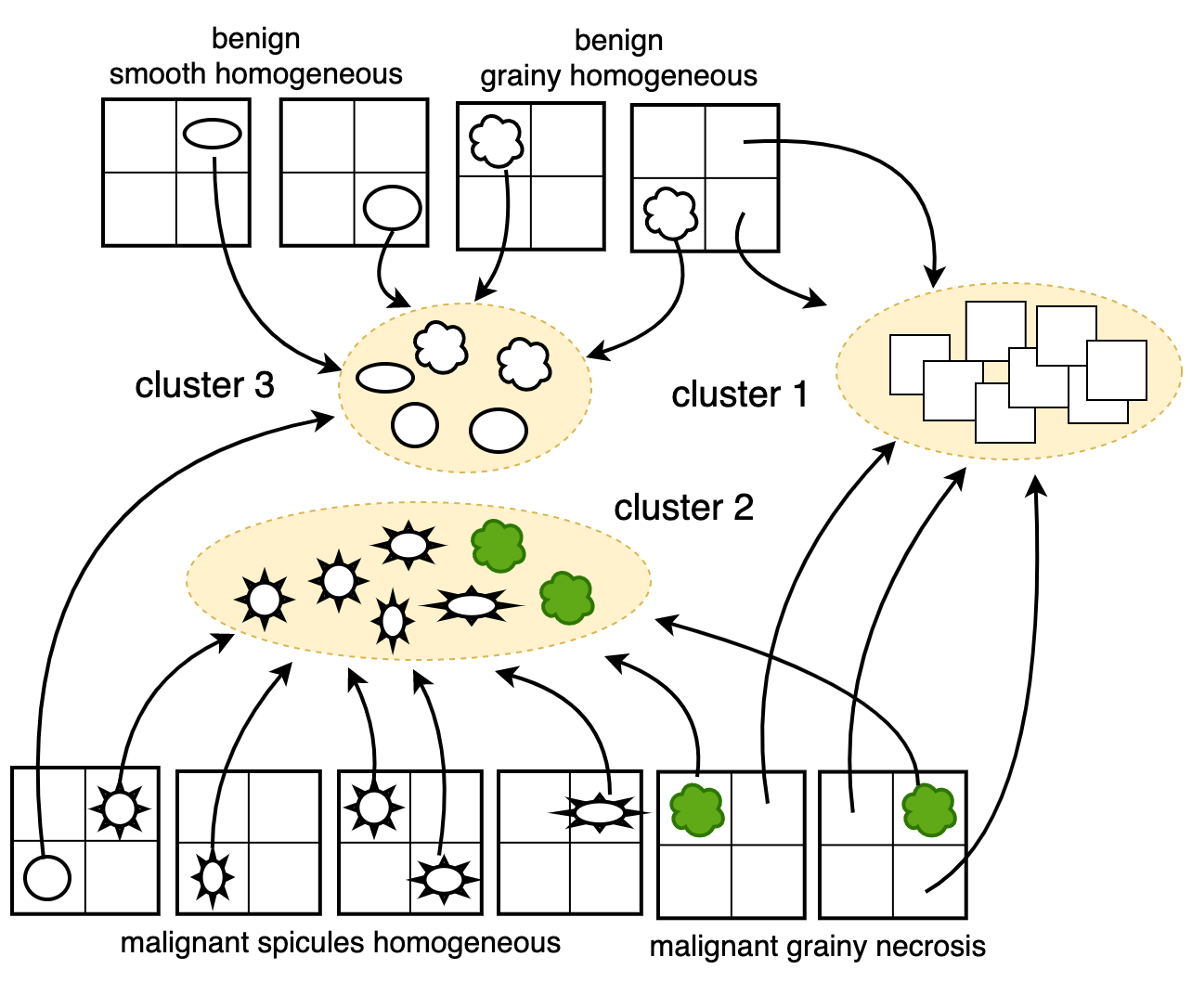}%
\caption{An illustration of the concept-based description of images consisting
of four patches and three clusters containing different patches}%
\label{f:concept_1_simple}%
\end{center}
\end{figure}

An example of the corresponding images and patches are schematically depicted
in Fig. \ref{f:concept_1_simple}. Prior probabilities $p(r,v)$ for all
concepts are shown in Table \ref{t:example_prior_pr}%

\begin{table}[tbp] \centering
\caption{Prior probabilities $p(r,v)$ for all concepts }%
\begin{tabular}
[c]{ccccccc}\hline
\multicolumn{2}{c}{$C^{(0)}$} & \multicolumn{3}{c}{$C^{(1)}$} &
\multicolumn{2}{c}{$C^{(2)}$}\\ \hline
$1$ & $2$ & $1$ & $2$ & $3$ & $1$ & $2$\\ \hline
$6/10$ & $4/10$ & $2/10$ & $4/10$ & $4/10$ & $8/10$ & $2/10$\\ \hline
\end{tabular}
\label{t:example_prior_pr}%
\end{table}%

Unconditional probabilities $p(l)$ are
\begin{equation}
p(1)=28/40,~p(2)=7/40,~p(3)=5/40.
\end{equation}

Table \ref{t:example_cond_pr} shows conditional probabilities $p(l\mid r,v)$.
Table \ref{t:example_poster_prob} contains posterior probabilities $p(r,v\mid
l)$ computed by using (\ref{Bayes_rule}). It follows from Table
\ref{t:example_poster_prob} that the second cluster contains patches which
show that the corresponding images are malignant with the probability $1$.
Moreover, the third cluster contains only patches with homogeneous nodules.%

\begin{table}[tbp] \centering
\caption{Conditional probabilities $p(l\mid r,v)$ }%
\begin{tabular}
[c]{cccccccc}\hline
& \multicolumn{2}{c}{$C^{(0)}$} & \multicolumn{3}{c}{$C^{(1)}$} &
\multicolumn{2}{c}{$C^{(2)}$}\\ \hline
$v$ & $1$ & $2$ & $1$ & $2$ & $3$ & $1$ & $2$\\ \hline
$K_{1}$ & $16/24$ & $3/4$ & $3/4$ & $6/8$ & $10/16$ & $22/32$ & $3/4$\\
$K_{2}$ & $7/24$ & $0$ & $0$ & $1/8$ & $5/16$ & $5/32$ & $1/4$\\
$K_{3}$ & $1/24$ & $1/4$ & $1/4$ & $1/8$ & $1/16$ & $5/32$ & $0$\\ \hline
\end{tabular}
\label{t:example_cond_pr}%
\end{table}%
%

\begin{table}[tbp] \centering
\caption{Posterior probabilities $p(r,v\mid l)$ }%
\begin{tabular}
[c]{cccccccc}\hline
& \multicolumn{2}{c}{$C^{(0)}$} & \multicolumn{3}{c}{$C^{(1)}$} &
\multicolumn{2}{c}{$C^{(2)}$}\\ \hline
$v$ & $1$ & $2$ & $1$ & $2$ & $3$ & $1$ & $2$\\ \hline
$K_{1}$ & $4/7$ & $3/7$ & $3/14$ & $6/14$ & $5/14$ & $11/14$ & $3/14$\\
$K_{2}$ & $1$ & $0$ & $0$ & $2/7$ & $5/7$ & $5/7$ & $2/7$\\
$K_{3}$ & $1/5$ & $4/5$ & $2/5$ & $2/5$ & $1/5$ & $1$ & $0$\\ \hline
\end{tabular}
\label{t:example_poster_prob}%
\end{table}%
%

\subsubsection{Inference}

Suppose that two embeddings of a new instance fall into the first cluster, and
two embedding falls into the third cluster. This implies that $\mathbf{(}%
s_{1},s_{2},s_{3}\mathbf{)}=(2,0,2)$. Probabilities $p(l\mid r,v)$ are taken
from Table \ref{t:example_cond_pr}. Hence, there holds for $C^{(r)}=v$:
\begin{align}
&  p(r,v\mid E_{1:R}^{\ast})\nonumber \\
&  =\frac{p(r,v)}{P\left \{  E_{1:R}^{\ast}\right \}  }\frac{4!}{2!\cdot
0!\cdot2!}p^{2}(1\mid r,v)\cdot p^{0}(2\mid r,v)\cdot p^{2}(2\mid r,v).
\end{align}

Probabilities $P\left \{  E_{1:R}^{\ast}\right \}  $ are computed from
(\ref{cond_prob}) for every $r=1,2,3$. They are $0.087$, $0.067$, $0.056$.
Final results are shown in Table \ref{t:new_example_prob}. It can be seen from
Table \ref{t:new_example_prob} that the tested image is
$<$%
benign%
$>$%
\ ($c^{(0)}=2$ with the probability $0.968$) with homogeneous nodules
($c^{(2)}=1$ with the probability $0.999$). The decision about the concept
\textbf{Contour} depends on the threshold $\gamma_{2}$. If $\gamma_{2}%
\leq0.630$, then we can say that nodules are smooth. If $\gamma_{2}>0.630$,
then this concept is not available or too uncertain to be included into the
image description. At the same time, the small probability of spicules implies
that the \textbf{Contour} is smooth or grainy with probability $0.945$.%

\begin{table}[tbp] \centering
\caption{Posterior probabilities $p(r,v\mid E_{1:R}^{\ast })$ }%
\begin{tabular}
[c]{cccccccc}\hline
& \multicolumn{2}{c}{$C^{(0)}$} & \multicolumn{3}{c}{$C^{(1)}$} &
\multicolumn{2}{c}{$C^{(2)}$}\\ \hline
$v$ & $1$ & $2$ & $1$ & $2$ & $3$ & $1$ & $2$\\ \hline
& $0.032$ & $0.968$ & $0.630$ & $0.315$ & $0.055$ & $0.999$ & $0.001$\\ \hline
\end{tabular}
\label{t:new_example_prob}%
\end{table}%

\subsection{Expert rules}

\subsubsection{Prior probabilities}

Let us consider the illustrative example under condition of using the expert
rule \textquotedblleft IF \textbf{Contour} is
$<$%
grainy%
$>$%
, THEN \textbf{Diagnosis} is
$<$%
malignant%
$>$%
\textquotedblright \ which\ can also be written as \textquotedblleft IF
$c^{(1)}=2$, THEN $c^{(0)}=1$\textquotedblright. The logical function
$g(\mathbf{c})$ corresponding to the rule is of the form:%
\begin{align}
g(\mathbf{c})  &  =[c^{(1)}=2]\rightarrow \lbrack c^{(0)}=1]\nonumber \\
&  =\lnot \left(  \lbrack c^{(1)}=2]\right)  \vee \lbrack c^{(0)}=1]\nonumber \\
&  =[c^{(0)}=1]\vee \lbrack c^{(1)}=1]\vee \lbrack c^{(1)}=3].
\label{expert_rule_1}%
\end{align}

The above rule is TRUE if at least one of the logical literals is TRUE.

Let us return to the above example. We have the following combinations
$\mathbf{z}$ of the concept labels (see Table \ref{t:example_descr_1}):
benign-smooth-homogeneous, benign-grainy-homogeneous,
malignant-spicules-homogeneous, malignant-grainy-necrosis. Let us find how the
prior probability $p(0,1)=P\left \{  C^{(0)}=1\right \}  $ will be changed when
the expert rule from the above example is available. The rule
\textquotedblleft IF \textbf{Contour} is
$<$%
grainy%
$>$%
, THEN \textbf{Diagnosis} is
$<$%
malignant%
$>$%
\textquotedblright \ is FALSE when $[c^{(0)}=2]\wedge \lbrack c^{(1)}=2]$, i.e.,
when the concept values are
$<$%
grainy%
$>$
and
$<$%
benign%
$>$%
. There are two images simultaneously having the concepts
$<$%
grainy%
$>$
and
$<$%
benign%
$>$%
. This implies that $g(\mathbf{z})=0$ for images with concepts
$<$%
benign-grainy-homogeneous%
$>$%
\ (see Fig. \ref{f:concept_1_simple}). Hence, we can write $\mathcal{C}%
_{0,1}=\{(1,3,1),(1,2,2)\}$, $\mathcal{C}=\{(1,3,1),(1,2,2),(2,1,1)\}$. Here
combinations with $g(\mathbf{z})=0$ are not provided for short. Probabilities
$P\{ \mathbf{C}=\mathbf{z}\}$ are shown in Table \ref{t:prior_prob_rule}.
Finally, we can write
\begin{equation}
U_{1}^{(0)}=\frac{P\{ \mathbf{C}=(1,3,1)\}+P\{ \mathbf{C}=(1,2,2)\}}{P\left \{
g(\mathbf{C})=1\right \}  },
\end{equation}
where
\begin{align}
P\left \{  g(\mathbf{C})=1\right \}   &  =P\{ \mathbf{C}=(1,3,1)\} \cdot
p(0,1)\nonumber \\
+P\{ \mathbf{C}  &  =(1,2,2)\} \cdot p(0,1)\nonumber \\
+P\{ \mathbf{C}  &  =(2,1,1)\} \cdot p(0,2),
\end{align}

and%
\begin{equation}
P\left \{  C^{(0)}=1\mid g(\mathbf{C})=1\right \}  =0.818.
\end{equation}

Let us find the prior probability $P\left \{  C^{(0)}=2\right \}  $. In this
case, there holds $\mathcal{C}_{0,1}=\{(2,1,1)\}$. Hence, we obtain
\begin{equation}
U_{2}^{(0)}=\frac{P\{ \mathbf{C}=(2,1,1)\}}{P\left \{  g(\mathbf{C})=1\right \}
},
\end{equation}
and
\begin{equation}
P\left \{  C^{(0)}=2\mid g(\mathbf{C})=1\right \}  =0.182.
\end{equation}

It is interesting to point out that the expert rule increases the prior
probability of the
$<$%
malignant%
$>$
and decreases the probability of the
$<$%
benign%
$>$%
. Indeed, two images with the concept values
$<$%
benign%
$>$
and
$<$%
grainy%
$>$
do not correspond to the expert rule. Therefore, these images can be regarded
as inadmissible for analyzing probabilities of the concept $c^{(0)}$. However,
it does not mean that they cannot be used for finding probabilities of other concepts.%

\begin{table}[tbp] \centering
\caption{Prior probabilities $P \{{\bf{C}}={\bf{z}}\}$}%
\begin{tabular}
[c]{cccc}\hline
$\mathbf{z}$ & $(1,3,1)$ & $(1,2,2)$ & $(2,1,1)$\\ \hline
$P\{ \mathbf{C}=\mathbf{z}\}$ & $4/10$ & $2/10$ & $2/10$\\ \hline
\end{tabular}
\label{t:prior_prob_rule}%
\end{table}%

\subsubsection{Updating posterior probabilities}

Let us again return to the above example and find the conditional
probabilities $P\left \{  E_{l}^{(0)}=v\mid g(\mathbf{C})=1\right \}  $,
$v=1,2$, for every cluster.

Cluster 1: The subset $\mathcal{C}_{0,1}(1)$ consists of combinations
$(1,3,1)$, $(1,2,2)$ because embeddings corresponding to all concepts are
included in $K_{1}$. The subset $\mathcal{C}_{0,2}(1)$ consists of the
combination $(2,1,1)$.

Cluster 2: The subset $\mathcal{C}_{0,1}(2)$ consists of combinations
$(1,3,1)$, $(1,2,2)$ because embeddings with the corresponding concept values
are included in $K_{2}$. The subset $\mathcal{C}_{0,2}(2)$ is empty.

Cluster 3: The subset $\mathcal{C}_{0,1}(3)$ consists of the combination
$(1,3,1)$ because one embedding from an image with the concept values
$(1,3,1)$ falls into $K_{3}$. The subset $\mathcal{C}_{0,2}(3)$ consists of
the combination $(2,1,1)$.

Finally, we can write using Tables \ref{t:example_cond_pr} and
\ref{t:prior_prob_rule}:
\begin{align}
V_{1}^{(0)}(1)  &  =\frac{P\{ \mathbf{C}=(1,3,1)\}+P\{ \mathbf{C}%
=(1,2,2)\}}{P\left \{  g(\mathbf{C})=1\right \}  }\nonumber \\
&  =\frac{0.6}{P\left \{  g(\mathbf{C})=1\right \}  },
\end{align}%
\begin{equation}
V_{1}^{(0)}(2)=V_{1}^{(0)}(1)=\frac{0.6}{P\left \{  g(\mathbf{C})=1\right \}  },
\end{equation}%
\begin{equation}
V_{1}^{(0)}(3)=\frac{P\{ \mathbf{C}=(1,3,1)\}}{P\left \{  g(\mathbf{C}%
)=1\right \}  }=\frac{0.4}{P\left \{  g(\mathbf{C})=1\right \}  },
\end{equation}
where
\begin{equation}
P\left \{  g(\mathbf{C})=1\right \}  =0.6\cdot16/24+0.6\cdot7/24+0.4\cdot
1/24=0.592.
\end{equation}

Hence, we obtain $V_{1}^{(0)}(1)=V_{1}^{(0)}(2)=\allowbreak1.\, \allowbreak
014$, $V_{1}^{(0)}(3)=\allowbreak0.676$. Similarly, we find
\begin{align}
V_{2}^{(0)}(1)  &  =V_{2}^{(0)}(3)=\frac{P\{ \mathbf{C}=(1,3,1)\}}{P\left \{
g(\mathbf{C})=1\right \}  }\nonumber \\
&  =\frac{0.4}{P\left \{  g(\mathbf{C})=1\right \}  },\ V_{2}^{(0)}(2)=0,
\end{align}
where
\begin{equation}
P\left \{  g(\mathbf{C})=1\right \}  =0.4\cdot3/4+0.0\cdot0+0.4\cdot
1/4=\allowbreak0.4.
\end{equation}

Hence, we obtain $V_{2}^{(0)}(1)=V_{2}^{(0)}(3)=\allowbreak1$, $V_{2}%
^{(0)}(2)=\allowbreak0$. Finally, we find the conditional probabilities
$P\left \{  E_{l}^{(r)}=v\mid g(\mathbf{C})=1\right \}  $ from
(\ref{cond_total_prob_3}), whose values are given in Table
\ref{t:example_cond_pr_updated}.%

\begin{table}[tbp] \centering
\caption{Conditional probabilities $P \left \{ E_{l}^{(r)}=v\mid g({\bf{C}})=1\right \} $ }%
\begin{tabular}
[c]{ccc}\hline
& \multicolumn{2}{c}{$c^{(0)}$}\\ \hline
$v$ & $1$ & $2$\\ \hline
$K_{1}$ & $\allowbreak0.676\,$ & $3/4$\\
$K_{2}$ & $\allowbreak0.296$ & $0$\\
$K_{3}$ & $0.028$ & $1/4$\\ \hline
\end{tabular}
\label{t:example_cond_pr_updated}%
\end{table}%

If we compare conditional probabilities of the concept $c^{(0)}$ obtained
without rules (see Table \ref{t:example_cond_pr}) and with rules (see Table
\ref{t:example_cond_pr_updated}), then we can see that the probabilities for
$c^{(0)}=1$ have been changed whereas probabilities for $c^{(0)}=2$ have not
been changed.

\section{Numerical experiments}

To study the proposed model, a synthetic dataset is constructed from the
well-known MNIST dataset \cite{LeCun-etal-98} which represents $28\times28$
pixel handwritten digit images. The original MNIST dataset has a training set
of $60,000$ instances and a test set of $10,000$ instances. The dataset is
available at http://yann.lecun.com/exdb/mnist/.%

\begin{figure}
[ptb]
\begin{center}
\includegraphics[
height=0.8717in,
width=3.5051in
]%
{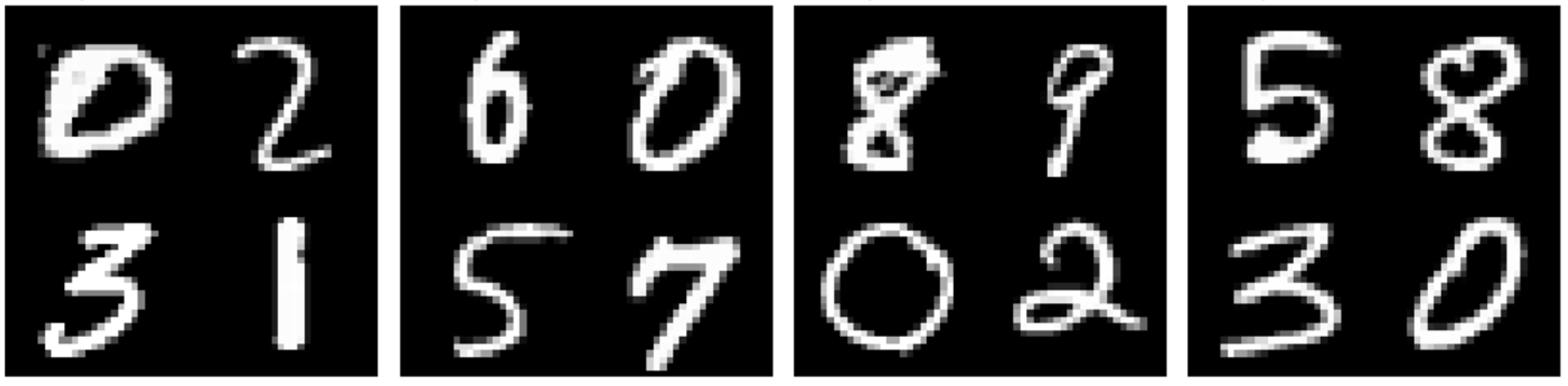}%
\caption{Examples of the modified MNIST dataset}%
\label{f:large_mnist}%
\end{center}
\end{figure}

Each instance in the synthetic dataset consists of four different digits
randomly taken from MNIST such that the instance has two digits in the first
row and two digits in the second row as it is shown in Fig.
\ref{f:large_mnist}. Each instance has the size $56\times56$. A similar
dataset is used in \cite{kim2023probabilistic}. Concepts are formed as follows:

\begin{itemize}
\item The concept $c^{(0)}$ (target) is defined as the largest digit among
four digits in each instance. This corresponds to the classification task with
seven classes (digits from $3$ till $9$) due to the difference of digits in
the instance.

\item Concepts $c^{(1)},...,c^{(10)}$ are binary and defined by the presence
of the corresponding number ($1,...,9,0$) in the instance.
\end{itemize}

The proposed model is compared with the original CBM \cite{koh2020concept}.
Preliminary numerical experiments have shown that the original CBM provides
outperforming results when the number of instances is large. Therefore, we
compare the proposed model with CBM by training models on small numbers of
instances (from $500$ till $5000$). The number of testing images is $20,000$.

Details of the proposed model are the following:

\begin{itemize}
\item Images are divided into 4 patches such that each patch contains one digit.

\item The autoencoder is constructed by using the convolution network, the
embedding size is $16$.

\item The expectation--maximization (EM) algorithm with $80$ clusters is used
for clustering.

\item The model is trained on numbers of epochs from $30$ till $50$ depending
on the number of instances taken for training.
\end{itemize}

CBM is also constructed on the convolution network which transforms the image
to an embedding. For each concept, a fully connected two-layer network
predicts the concept value. The cross-entropy loss function is used. A
modification of CBM with the joint training of the bottleneck and targets is
used such that the joint bottleneck minimizes the weighted sum of loss
functions with coefficients 1. The F1 measure is used as an accuracy measure
in experiments because the training set is imbalanced due to the considered
structure of concepts and images.

\begin{figure}
[ptb]
\begin{center}
\includegraphics[
height=3.7187in,
width=4.9492in
]%
{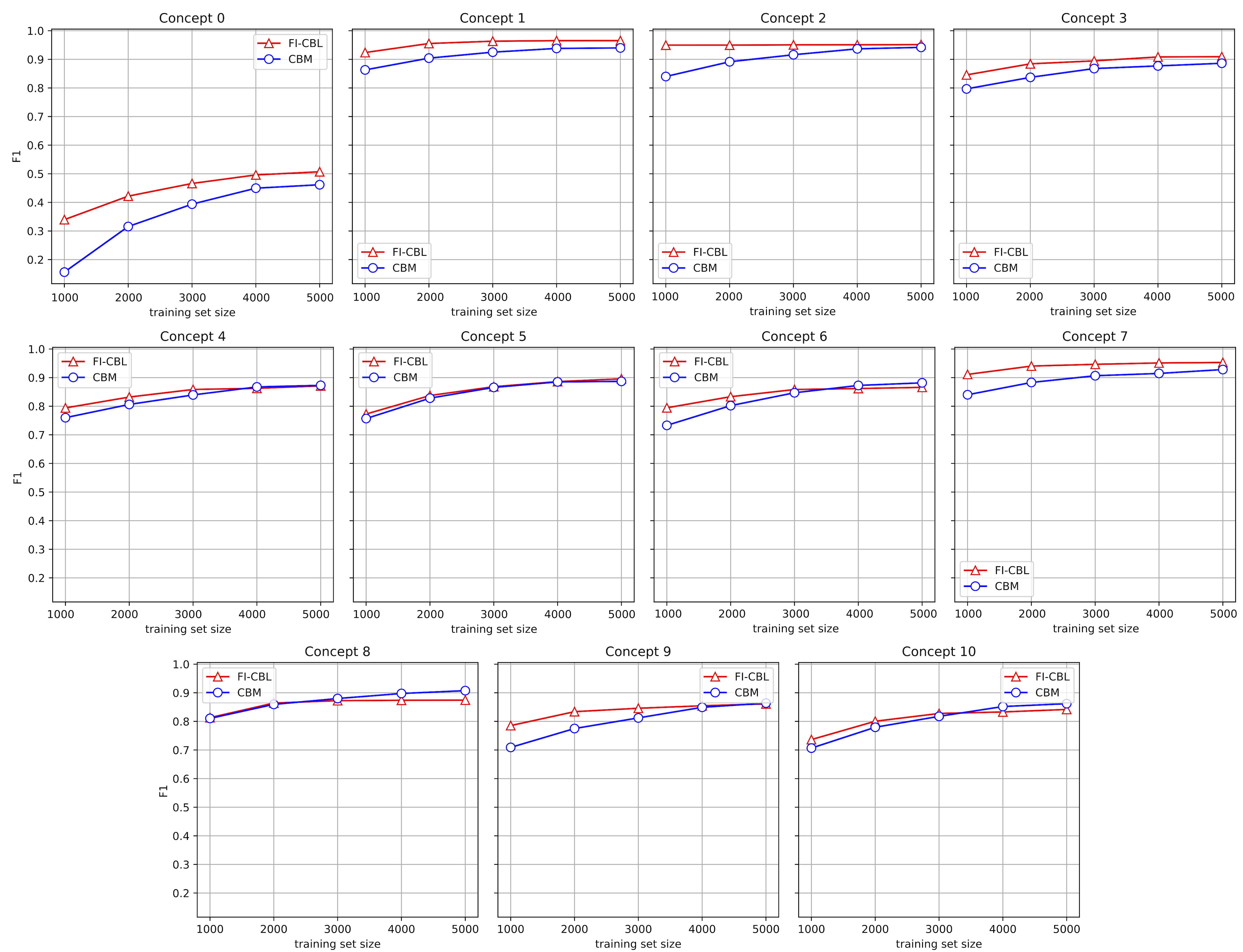}%
\caption{The F1 measures of the proposed method and CBM as functions of the
training set size for all concepts of the modified MNIST dataset}%
\label{f:exp5}%
\end{center}
\end{figure}

F1 measures as functions of the training set size for all concepts of the
modified MNIST dataset obtained by using the proposed method and CBM are shown
in Fig. \ref{f:exp5}. It can be seen from Fig. \ref{f:exp5} that FI-CBL
outperforms the original CBM when the training set size is small (smaller than
$5000$ instances). This can be explained by the fact that the neural network
implementing CBM requires a significantly larger number of instances for
training. On the contrary, FI-CBL allows us to obtain acceptable results with
a small number of training instances. At the same time, CBM becomes better as
the training set increases. In this case, FI-CBL requires finer tuning of the
number of clusters and the size of embeddings.%

\begin{figure}
[ptb]
\begin{center}
\includegraphics[
height=2.2883in,
width=3.0536in
]%
{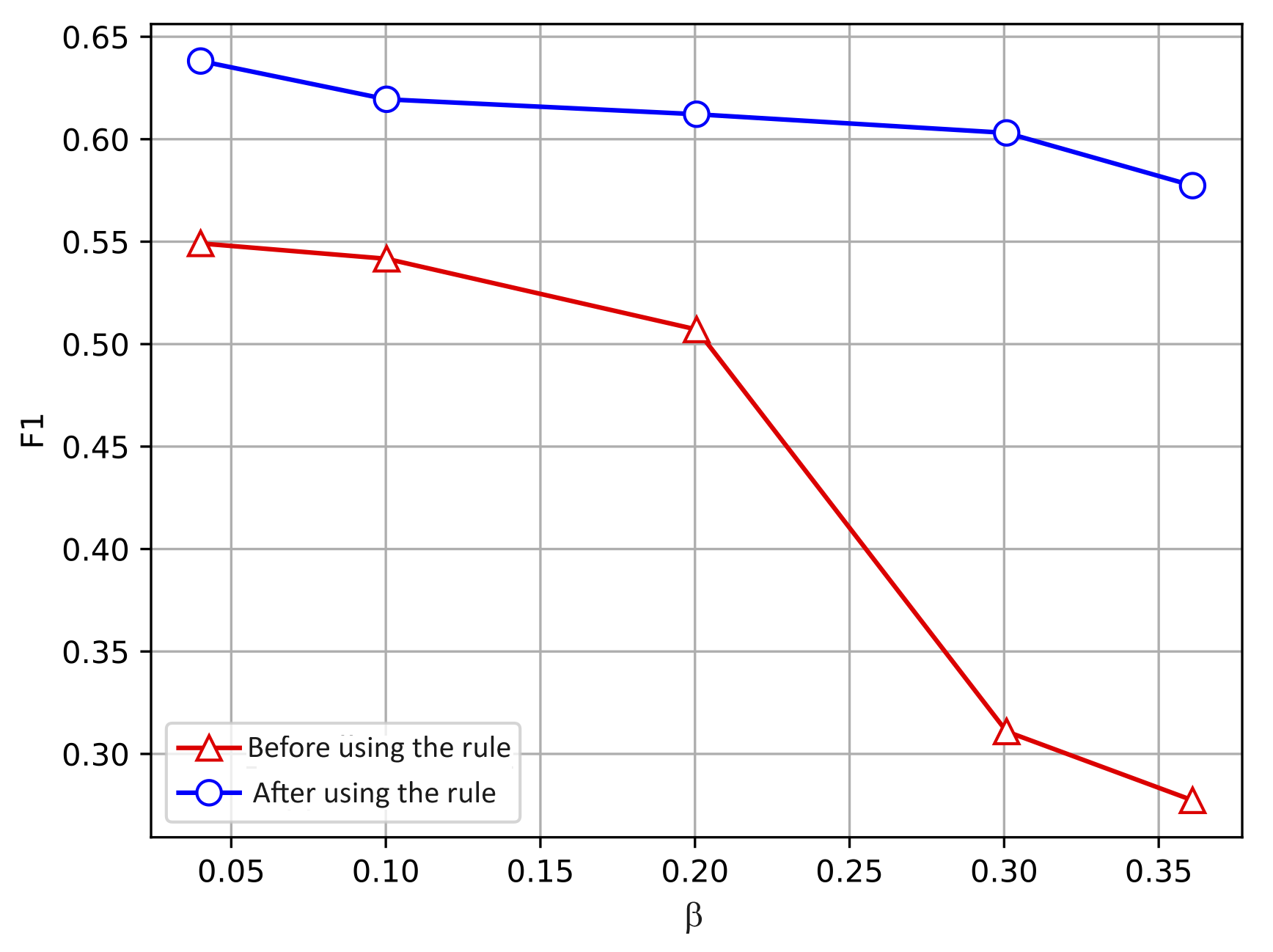}%
\caption{F1-measures as functions of $\beta$ for cases when any rule is not
used (the curve with triangle markers), and when the rule \textquotedblleft IF
$c^{(9)}=1$, THEN $c^{(0)}=1$\textquotedblright \ is used (the curve with
circle markers) for the modified MNIST}%
\label{f:mnist_4_f1_rule}%
\end{center}
\end{figure}

To study how an expert rule impacts on the prediction accuracy, we invert some
correct labels that satisfy the expert rule. The rule is \textquotedblleft IF
$c^{(9)}=1$, THEN $c^{(0)}=1$\textquotedblright. This is an obvious rule which
means that if there is the digit $9$ among four digits in an instance, then
the target is $9$. Let $\beta$ be the portion of instances in the training
set, whose labels are inverted. Fig. \ref{f:mnist_4_f1_rule} illustrates how
F1 measures depend on values of $\beta$ for two cases: before using the rule
(the curve with triangle markers) and after using the rule (the curve with
circle markers). It can be seen from Fig. \ref{f:mnist_4_f1_rule} that the use
of the expert rule allows us to significantly correct the model and to obtain
better predictions (the corresponding function slowly decreases with
increase). At the same time, if the rule is not used, then the F1 measure
quickly decreases with increase of $\beta$.%

We also consider a modification of the Large-scale CelebFaces Attributes
(CelebA) dataset \cite{liu2015deep}. The original dataset contains $200,000$
images, each annotated with $40$ face attributes and $10,000$ classes. The
modification is restricted by $20$ classes such that images with labels from
$1$ till $500$ have the label $0$, images with labels from $501$ till $1000$
have the label $1$, etc. The number of concepts is $40$. All concepts are
binary. To compare FI-CBL with CBM, we find the F1 measure averaged over all
$40$ concepts. The number of testing images is approximately $65,000$, i.e.,
40\% of the original dataset. The K-means algorithm with $256$ clusters is
used for clustering. The embedding size is $32$. The number of patches is
$35$. To ensure that the whole concepts fall into separate patches, we propose
using overlapping patches similar to sliding windows in convolutional neural networks.

\begin{figure}
[ptb]
\begin{center}
\includegraphics[
height=2.3984in,
width=3.1751in
]%
{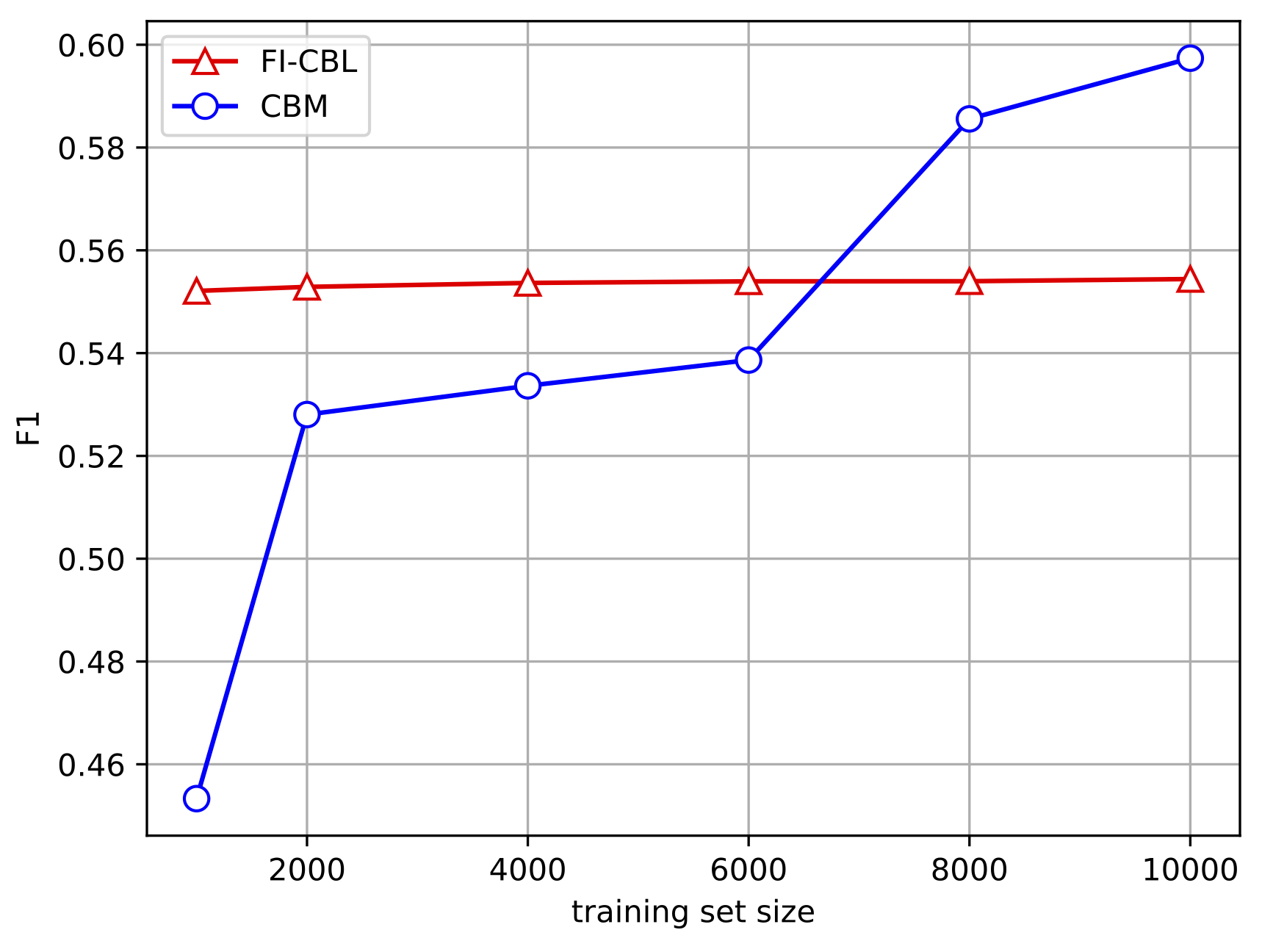}%
\caption{The averaged F1 measures as functions of the training set size for
the modified CelebA dataset obtained by using FI-CBL and CBM}%
\label{f:celeba_f1}%
\end{center}
\end{figure}

F1 measures as functions of the training set size of the modified CelebA
dataset obtained by using FI-CBL and CBM are shown in Fig. \ref{f:celeba_f1}.
Similarly to the previous numerical example with the MNIST\ dataset, FI-CBL
outperforms the original CBM when the training set size is small (smaller than
$4000$ instances).%

Let us consider another numerical example with the original MNIST dataset. All
instances are annotated by the following concepts:

0 - target, $\mathcal{C}^{(0)}=\{0,1\}$, values of the target are given in
(\ref{small_MNIST_concepts});

1 - the even/odd digit, $\mathcal{C}^{(1)}=\{0,1\}$;

2 - the digit is not less than 5 / is less than 5, $\mathcal{C}^{(2)}=\{0,1\}$;

3 - the digit is a remainder of division by 3, $\mathcal{C}^{(3)}=\{0,1,2\}$.

In sum, digits are represented by the following sets of concepts:%
\begin{align}
0  &  \rightarrow \lbrack0,0,1,0];\ 1\rightarrow \lbrack1,1,1,1];\ 2\rightarrow
\lbrack0,0,1,2];\nonumber \\
3  &  \rightarrow \lbrack1,1,1,0];\ 4\rightarrow \lbrack0,0,1,1];\ 5\rightarrow
\lbrack0,1,0,2];\nonumber \\
6  &  \rightarrow \lbrack1,0,0,0];\ 7\rightarrow \lbrack0,1,0,1];\ 8\rightarrow
\lbrack1,0,0,2];\nonumber \\
9  &  \rightarrow \lbrack0,1,0,0]. \label{small_MNIST_concepts}%
\end{align}

The EM algorithm with $128$ clusters is used for clustering. The embedding
size is $16$. The testing set consists of $24,000$ images.

\begin{figure}
[ptb]
\begin{center}
\includegraphics[
height=3.0244in,
width=3.0244in
]%
{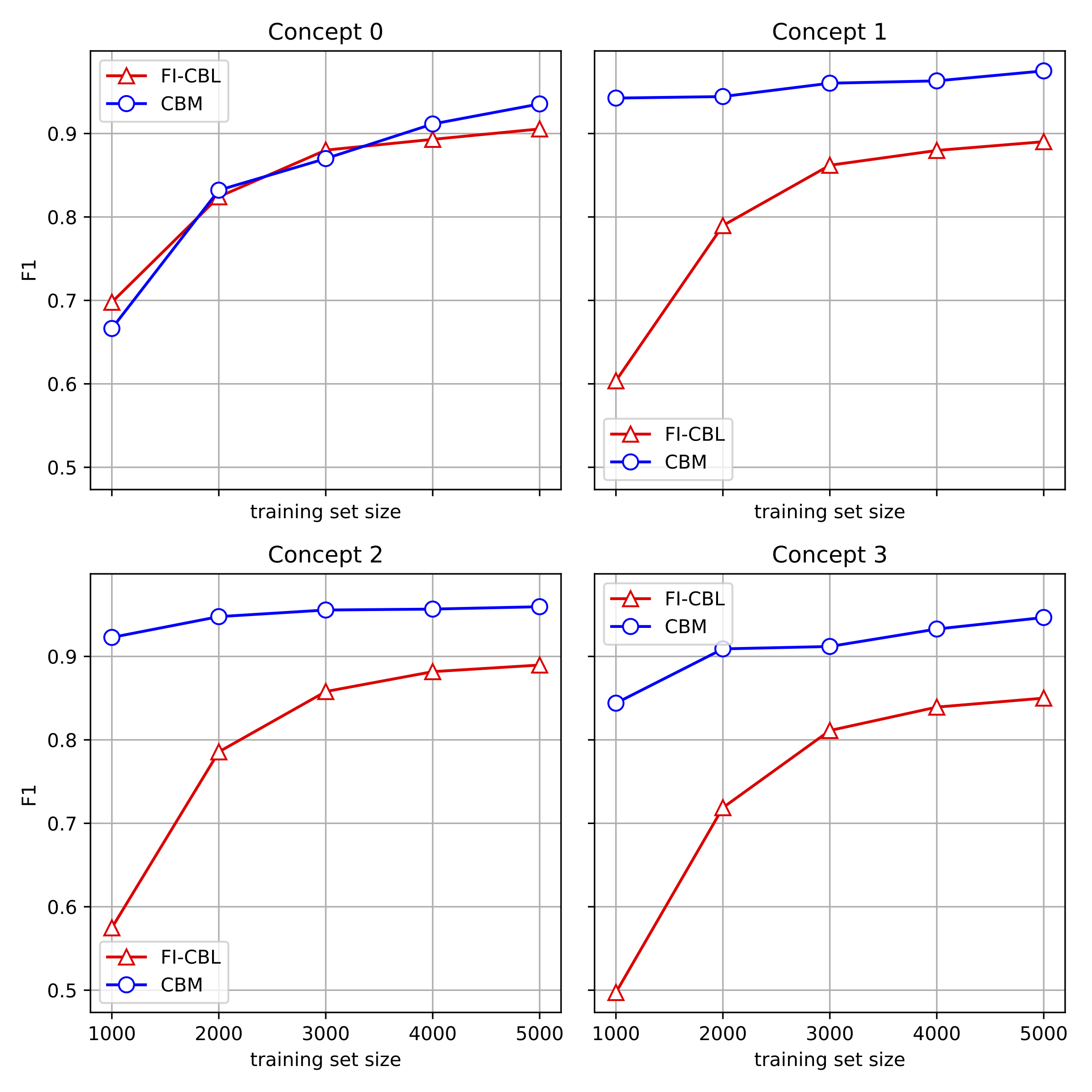}%
\caption{The F1 measures of the proposed method and CBM as functions of the
training set size for all concepts of the original MNIST dataset}%
\label{f:mnist_1_f1}%
\end{center}
\end{figure}

\begin{figure}
[ptb]
\begin{center}
\includegraphics[
height=2.2543in,
width=3.0532in
]%
{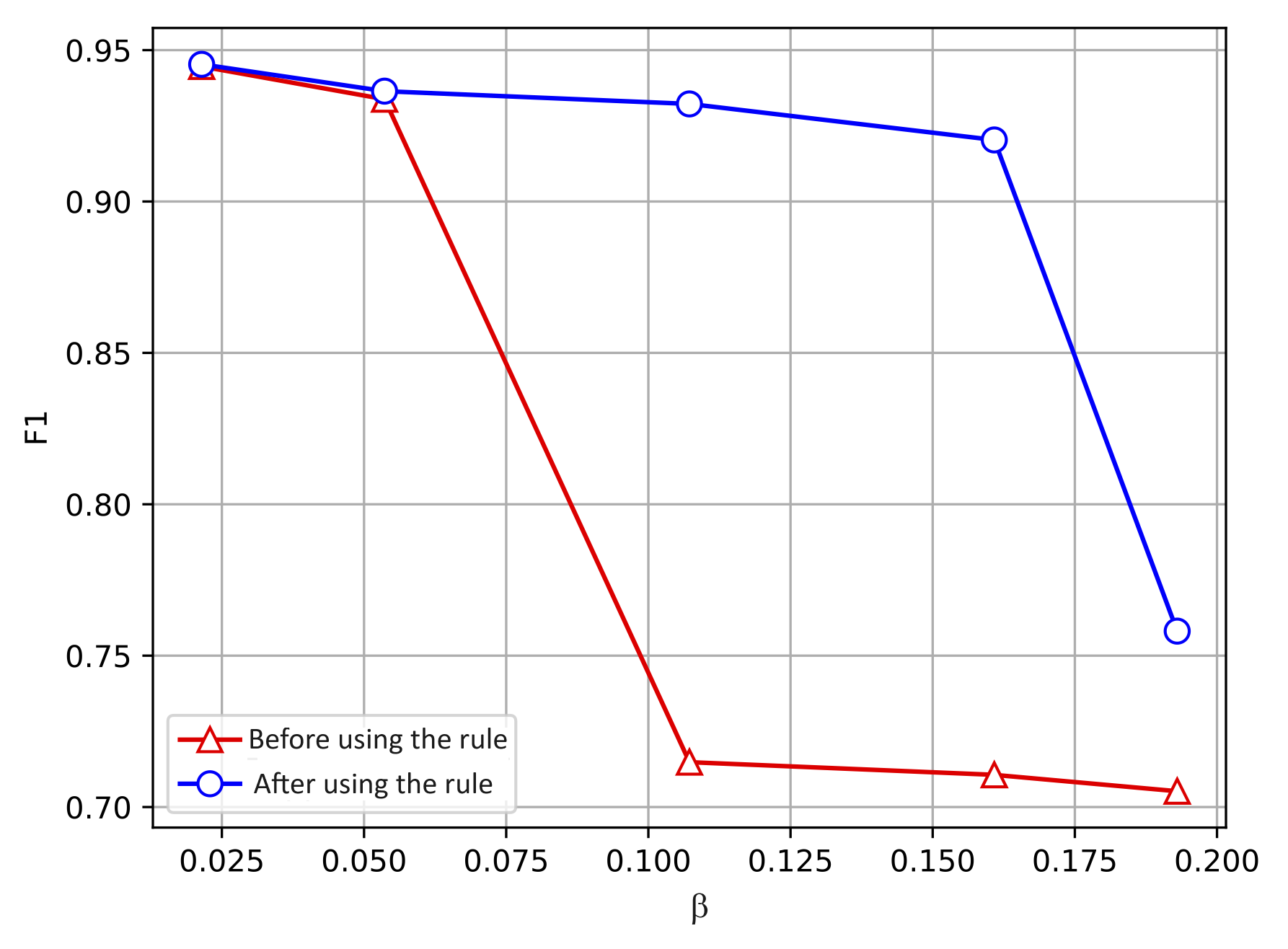}%
\caption{F1 measures as functions of $\beta$ for two cases: before using the
rule and after using the rule for the MNIST\ dataset}%
\label{f:violation_f1}%
\end{center}
\end{figure}

F1 measures as functions of the training set size for four concepts of the
original MNIST dataset obtained by using the proposed method and CBM are shown
in Fig. \ref{f:mnist_1_f1}. One can see from Fig. \ref{f:mnist_1_f1} that CBM
outperforms FI-CBL in most cases even when the training set size is small.
This numerical example aims to show that CBM can provide better results even
when the number of training instances is small. At the same time, the next
numerical experiments aim to study how expert rules can correct incorrect
concepts and provide better results in comparison with CBM. We aim to show now
that CBM provides worse results in comparison with FI-CBL when labels are
noisy. We again invert some correct labels that satisfy the expert rule. The
rule is \textquotedblleft IF $c^{(1)}=1$ AND $c^{(2)}=1$, THEN $c^{(0)}%
=1$\textquotedblright. Fig. \ref{f:violation_f1} illustrates the difference
between F1 measures as functions of $\beta$ for two cases: before using the
rule and after using the rule. It can be seen from Fig. \ref{f:violation_f1}
that the use of the rule allows us to significantly correct the model and to
obtain better predictions. When values of $\beta$ exceed $0.16$, then both the
models show worse results.%

%

In the next experiment, we invert half of the target labels, i.e., the target
is completely confused. We aim to improve the model prediction by
incorporating the increasingly detailed expert rules:

\begin{itemize}
\item $g_{1}(\mathbf{c})=\left[  c^{(1)}=1\wedge c^{(2)}=1\rightarrow
c^{(0)}=1\right]  $;

\item $g_{2}(\mathbf{c})=\left[  \left(  c^{(1)}=0\wedge c^{(2)}=1\rightarrow
c^{(0)}=0\right)  \wedge \left(  c^{(1)}=1\wedge c^{(2)}=1\rightarrow
c^{(0)}=1\right)  \right]  $;

\item $g_{3}(\mathbf{c})=\left[  (c^{(1)}=0\wedge c^{(2)}=0)\vee
(c^{(1)}=1\wedge c^{(2)}=1)\leftrightarrow c^{(0)}=1\right]  $.
\end{itemize}

\begin{figure}
[ptb]
\begin{center}
\includegraphics[
height=1.625in,
width=4.843in
]%
{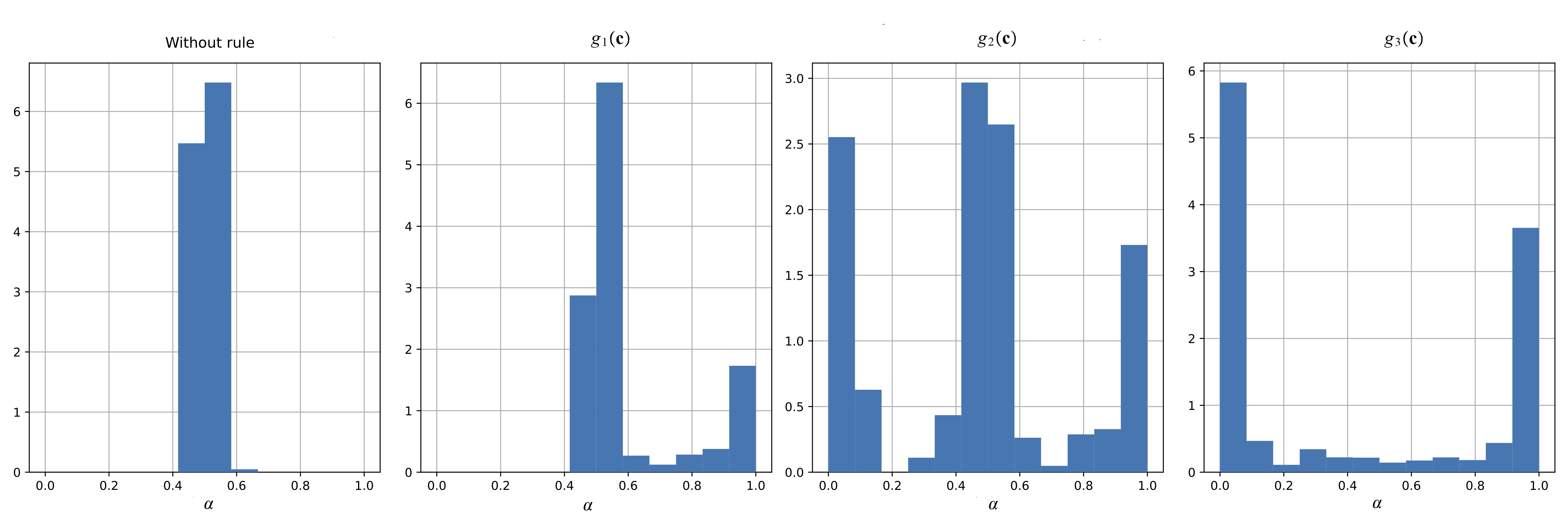}%
\caption{Four histograms of the target prediction probabilities obtained under
conditions of confused concept labels and different expert rules for the
original MNIST dataset}%
\label{f:density}%
\end{center}
\end{figure}

Fig. \ref{f:density} shows histograms of the binary target prediction
probabilities $\alpha$ without rules and with different rules. It can be seen
from the first histogram that most prediction probabilities are close to
$0.5$. This implies that the inversion of the target labels leads to the total
uncertainty, i.e. the model cannot correctly classify instances. The first
rule $g_{1}(\mathbf{c})$ partially improves results. One can see from the
second histogram that a small part of instances are classified with
probabilities close to $1$. However, so far most prediction probabilities are
close to $0.5$ despite the rule. The second rule $g_{2}(\mathbf{c})$ consists
of two rules and corrects many inverted labels. One can see from the third
histogram that the uncertainty of the target probabilities significantly
decreases in comparison with the previous cases. Finally, the third rule
$g_{3}(\mathbf{c})$, which is the most strong one due to the logical operation
\textquotedblleft if and only if\textquotedblright \ denoted as
\textquotedblleft$\leftrightarrow$\textquotedblright, provides the best
results shown in the fourth histogram. We can see that the uncertainty of
predictions is minimal. This is a very important observation which illustrates
how the incorporated expert rules are able to improve the model.%

\section{Conclusion}

Let us point out advantages and disadvantages of FI-CBL.

\emph{Advantages}:

\begin{itemize}
\item FI-CBL is transparent. In contrast to the neural network implementation
of CBL where the network as well as the bottleneck layer are a black box, the
proposed model has a clear and explicit sequence of calculations. All
probabilities (conditional and unconditional) have a clear frequency
interpretation. It allows us to extend the model in order to take into account
the possible limitation of instances in datasets and to apply robust
statistical methods which can correct the posterior probabilities and improve
the classification. Moreover, one can also observe entire processes of
training and inference in order to understand results of modeling.

\item FI-CBL is flexible. We can simply change the number of patches, the
number of clusters, the autoencoder architecture, threshold of decision making.

\item FI-CBL provides outperforming results when the number of training
instances is small.

\item FI-CBL is interpretable.
\end{itemize}

\emph{Disadvantages}:

\begin{itemize}
\item The main disadvantage is to implement a perfect clusterization and to
guess a proper number of clusters. We could learn the clusterization procedure
jointly with the autoencoder, but this approach may significantly complicate
the method and reduce its positive property to deal with small datasets.

\item The effectiveness of FI-CBL greatly depends on the size of patches. A
large difference in the size of concepts in an image can lead to significant
deterioration of the model. One of the ways to overcome this problem is to use
sliding windows for producing patches as it has been implemented in the
numerical experiment with CelebA dataset. However, this way requires
additional studies.

\item We need to store the dataset in order to use it every time when a new
instance is analyzed. On the other hand, we can store only a matrix of all
conditional probabilities $P\left \{  E_{l}^{(r)}=v\mid g(\mathbf{C}%
)=1\right \}  $ if the set of expert rules is not changed. If many clusters and
expert rules are implemented, then the matrix can be very large.

\item Experiments have illustrated that FI-CBL may be inferior to CBMs when
the number of training data is large.
\end{itemize}

The above disadvantages can be regarded as problems whose solutions are
direction for further research. In addition, the probabilistic approach used
in FI-CBL allows us to simply extend the method to solve several problems of
machine learning, including anomaly detection, unlearning, the attention
mechanism, etc. These are also directions for further research.

\bibliographystyle{unsrt}
\bibliography{Autoencoder,Classif_bib,Cluster_bib,Concept,Explain,Explain_med,MIL,MYBIB}

\end{document}